\documentclass{article}

\usepackage[final,nonatbib]{nips_2017}

\usepackage[utf8]{inputenc} 
\usepackage[T1]{fontenc}    
\usepackage{hyperref}       
\usepackage{url}            
\usepackage{booktabs}       
\usepackage{amsfonts}       
\usepackage{nicefrac}       
\usepackage{microtype}      
\usepackage[square,sort,comma,numbers]{natbib}
\usepackage{amsmath}
\usepackage{graphicx}

\title{Softmax GAN}

\author{
  Min Lin \\
  Qihoo 360 Technology co. ltd\\
  Beijing, China, 100871 \\
  \texttt{mavenlin@gmail.com} \\
}

\begin{document}

\maketitle

\begin{abstract}
  Softmax GAN is a novel variant of Generative Adversarial Network
  (GAN). The key idea of Softmax GAN is to replace the classification
  loss in the original GAN with a softmax cross-entropy loss in the
  sample space of one single batch. In the adversarial learning of $N$
  real training samples and $M$ generated samples, the target of
  discriminator training is to distribute all the probability mass to
  the real samples, each with probability $\frac{1}{M}$, and
  distribute zero probability to generated data. In the generator
  training phase, the target is to assign equal probability to all
  data points in the batch, each with probability
  $\frac{1}{M+N}$. While the original GAN is closely related to Noise
  Contrastive Estimation (NCE), we show that Softmax GAN is the
  Importance Sampling version of GAN. We futher demonstrate with
  experiments that this simple change stabilizes GAN training.
\end{abstract}

\section{Introduction}
Generative Adversarial Networks(GAN) \cite{GAN} has achieved great
success due to its ability to generate realistic samples. GAN is
composed of one Discriminator and one Generator. The discriminator
tries to distinguish real samples from generated samples, while the
generator counterfeits real samples using information from the
discriminator.  GAN is unique from the many other generative
models. Instead of explicitly sampling from a probability
distribution, GAN uses a deep neural network as a direct generator
that generates samples from random noises. GAN has been proved to work
well on several realistic tasks, e.g. image inpainting, debluring and
imitation learning.

Despite its success in many applications, GAN is highly unstable in
training. Careful selection of hyperparameters is often necessary to
make the training process converge \cite{DCGAN}. It is often believed
that this instability is caused by unbalanced discriminator and
generator training. As the discriminator utilizes a logistic loss, it
saturates quickly and its gradient vanishes if the generated samples
are easy to separate from the real ones. When the discriminator fails
to provide gradient, the generator stops updating. Softmax GAN
overcomes this problem by utilizing the softmax cross-entropy loss,
whose gradient is always non-zero unless the softmaxed distribution
matches the target distribution.

\section{Related Works}
There are many works related to improving the stability of GAN
training. DCGAN proposed by Radford et. al. \cite{DCGAN} comes up with
several empirical techniques that works well, including how to apply
batch normalization, how the input should be normalized, and which
activation function to use. Some more techniques are proposed by
Salimans et. al. \cite{salimans2016improved}. One of them is minibatch
discrimination. The idea is to introduce a layer that operates across
samples to introduce coordination between gradients from different
samples in a minibatch. In this work, we achieve a similar effect
using softmax across the samples. We argue that softmax is more
natural and explanable and yet does not require extra parameters.

Nowozin et. al. \cite{fgan} generalizes the GAN training loss from
\textit{Jensen-Shannon divergence} to any f-divergence
function. Wasserstein distance is mentioned as a member of another
class of probability metric in this paper but is not
implemented. Under the f-GAN framework, training objectives with more
stable gradients can be developed. For example, the Least Square GAN
\cite{mao2016least} uses $l_2$ loss function as the objective, which
achieves faster training and improves generation quality.

Arjovsky et. al. managed to use Wasserstein distance as the objective
in their Wasserstein GAN (WGAN) \cite{WGAN} work. This new objective
has non-zero gradients everywhere. The implementation is as simple as
removing the sigmoid function in the objective and adding weight
clipping to the discriminator network. WGAN is shown to be free of the
many problems in the original GAN, such as mode collapse and unstable
training process. A related work to WGAN is Loss-Sensitive GAN
\cite{LSGAN}, whose objective is to minimize the loss for real data
and maximize it for the fake ones. The common property of Least Square
GAN, WGAN, Loss-Sensitive GAN and this work is the usage of objective
functions with non-vanishing gradients.

\section{Softmax GAN}

We denote the minibatch sampled from the training data and the
generated data as $B_+$ and $B_-$ respectively. $B=B_++B_-$ is the
union of $B_+$ and $B_-$. The output of the discriminator is
represented by $\mu^\theta(x)$ parameterized by $\theta$.
$Z_B=\sum_{x\in{B}}{e^{-\mu^\theta(x)}}$ is the partition function of
the softmax within batch $B$. We use $x$ for samples from $B_+$ and
$x'$ for generated samples in $B_-$. As in GAN, generated samples are
not directly sampled from a distribution. Instead, they are generated
directly from a random variable $z$ with a trainable generator
$G^\psi$.

We softmax normalized the energy of the all data points within $B$,
and use the cross-entropy loss for both the discriminator and the
generator. The target of the discriminator is to assign the
probability mass equally to all samples in $B_+$, leaving samples in
$B_-$ with zero probability.

\begin{equation}
  t_D(x)=
  \begin{cases}
    \frac{1}{|B_+|}, & \text{if}\ x\in{B_+} \\
    0, & \text{if}\ x\in{B_-} \\
  \end{cases}
\end{equation}

\begin{align}
  L_D &= - \sum_{x\in{B}}t_D(x)\ln{\frac{e^{-\mu^\theta(x)}}{Z_B}} \nonumber \\
  &= - \sum_{x\in{B_+}}{
    \frac{1}{|B_+|}\ln{\frac{e^{-\mu^\theta(x)}}{Z_B}}
  } - \sum_{x'\in{B_-}}{
    0\ln{\frac{e^{-\mu^\theta(x')}}{Z_B}}
  } \nonumber \\
  &= \sum_{x\in{B_+}}{
    \frac{1}{|B_+|}\mu^\theta(x)
  } + \ln{Z_B}
  \label{eqn:Ld}
\end{align}

For generator, the target is to assign the probability mass equally to
all the samples in $B$.

\begin{equation}
  t_G(x) = \frac{1}{|B|}
\end{equation}

\begin{align}
  \label{eqn:Lg}
  L_G &= - \sum_{x\in{B}}t_G(x)\ln{\frac{e^{-\mu^\theta(x)}}{Z_B}} \nonumber \\
  &= - \sum_{x\in{B_+}}{
    \frac{1}{|B|}\ln{\frac{e^{-\mu^\theta(x)}}{Z_B}}
  } - \sum_{x'\in{B_-}}{
    \frac{1}{|B|}\ln{\frac{e^{-\mu^\theta(x')}}{Z_B}}
  } \nonumber \\
  &= \sum_{x\in{B_+}}{
    \frac{1}{|B|}\mu^\theta(x)
  } + \sum_{x'\in{B_-}}{
    \frac{1}{|B|}\mu^\theta(x')
  } + \ln{Z_B}
\end{align}

\section{Relationship to Importance Sampling}

It has been pointed out in the original GAN paper that GAN is similar
to NCE \cite{NCE} in that both of them use a binary logistic
classification loss as the surrogate function for training of a
generative model. GAN improves over NCE by using a trained generator
for noise samples instead of fixing the noise distribution. In the
same sense as GAN is related to NCE
\cite{goodfellow2014distinguishability}, this work can be seen as the
Importance Sampling version of GAN \cite{IS}. We prove it as follows.

\subsection{Discriminator}

We use $\xi^\phi(x)$ to represent energy function and
$Z=\int_{x}{e^{-\xi^\phi(x)}}$ is the partition function. The
probability density function is then
$p^\phi(x)=\frac{e^{-\xi^\phi(x)}}{Z}$. With $O$ as the observed
training example, the maximum likelyhood estimation loss function is
as follows
\begin{equation}
  J = \frac{1}{|O|}\sum_{x\in{O}}\xi^\phi(x) + \log{\int_{x'}{e^{-\xi^\phi(x')}}}dx
\end{equation}
\begin{equation}
  \nabla_{\phi}{J} = \frac{1}{|O|}\sum_{x\in{O}}\nabla_{\phi}\xi^\phi(x)
  - \mathbb{E}_{x'\sim{p^\phi(x')}}{\nabla_{\phi}{\xi^\phi(x')}}
\end{equation}

As it is usually difficult to sample from $p^\phi$, Importance
Sampling instead introduces a known distribution $q(x)$ to sample
from, resulting in:

\begin{equation}
  \nabla_{\phi}{J} = \frac{1}{|O|}\sum_{x\in{O}}\nabla_{\phi}\xi^\phi(x)
  - \mathbb{E}_{x'\sim{q(x')}}{\frac{p^\phi(x')}{q(x')}\nabla_{\phi}{\xi^\phi(x')}}
\end{equation}

In the biased Importance Sampling \cite{bengio2008adaptive}, the above
is converted to the following biased estimation which can be
calculated without knowing $p^\phi$:

\begin{equation}
  \label{eqn:estimatedGradient}
  \widehat{\nabla_\phi{J}} = \frac{1}{|B_+|}\sum_{x\in{B_+}}\nabla_\phi\xi^\phi(x)
  - \frac{1}{R}\sum_{x'\in{Q}}{r(x')\nabla_{\phi}{\xi^\phi(x')}}
\end{equation}

where $r(x') = \frac{e^{-\xi^\phi(x')}}{q(x')}$,
$R=\sum_{x'\in{Q}}r(x')$. And $Q$ is a batch of data sampled from $q$.
At this point, we reparameterize $e^{-\xi^\phi(x)}=e^{-\mu^\theta(x)}q(x)$.

\begin{equation}
  \label{eqn:reparam}
  \widehat{\nabla_\theta{J}} = \frac{1}{|B_+|}\sum_{x\in{B_+}}\nabla_\theta\mu^\theta(x)
  - \frac{1}{\sum_{y\in{Q}}{e^{-\mu^\theta(y)}}}\sum_{x'\in{Q}}{e^{-\mu^\theta(x')}\nabla_{\theta}{\mu^\theta(x')}}
\end{equation}

Without loss of generality, we assume $|B_+|=|B_-|$ and replace $Q$
with $B=B_++B_-$ in equation \ref{eqn:reparam}, namely
$q(x)=\frac{1}{2}p_{data}(x) + \frac{1}{2}p_G(x)$, and compare the
above with equation \ref{eqn:Ld}. It is easy to see that the above is
the gradient of $L_D$. In other words, the discriminator loss function
in Softmax GAN is performing maximum likelihood on the observed real
data with Importance Sampling to estimate the partition function.

With infinite number of real samples, the optimal solution is

\begin{equation}
  \label{eqn:discriminator_result}
  e^{-\mu^\theta(x)}=C\frac{p_D}{\frac{p_D+p_G}{2}}
\end{equation}

$C$ is a constant.

\subsection{Generator}
We substitute equation \ref{eqn:discriminator_result} into
\ref{eqn:Lg}. The \textit{lhs} of equation \ref{eqn:Lg} gives

\begin{equation}
  -\sum_{x\in{B}}\frac{1}{|B|}\ln{\frac{p_D}{\frac{p_D+p_G}{2}}}
  - \ln{C} = KL(\frac{p_D+p_G}{2}\|p_D)
\end{equation}

The gradient of the \textit{rhs} can be seen as biased Importance
Sampling as well,

\begin{equation}
  \frac{-\sum_{x\in{B}}\frac{p_D}{\frac{p_D+p_G}{2}}
    \frac{\nabla{p_G}}{p_D+p_G}}{\sum_{x\in{B}}\frac{p_D}{\frac{p_D+p_G}{2}}}
  \approx -\mathbb{E}_{x\sim{p_D}}{\frac{\nabla{p_G}}{p_D+p_G}}
\end{equation}

which optimizes $-\mathbb{E}_{x\sim{p_D}}{\ln(p_D+p_G)} =
KL(p_D\|\frac{p_D+p_G}{2}) - \mathbb{E}_{x\sim{p_D}}\ln{2p_D}$. After
removing the constants, we get

\begin{equation}
L_G = KL(\frac{p_D+p_G}{2}\|p_D) + KL(p_D\|\frac{p_D+p_G}{2})
\end{equation}

Thus optimizing the objective of the generator is equivalent to
minimizing the \textit{Jensen-Shannon divergence} between $p_D$ and
$\frac{p_D+p_G}{2}$ with Importance Sampling.

\subsection{Importance Sampling's link to NCE}
\begin{figure}
  \caption{Relationship of Algorithms}
  \label{fig:comparechart}
  \centering
  \includegraphics[width=0.6\textwidth]{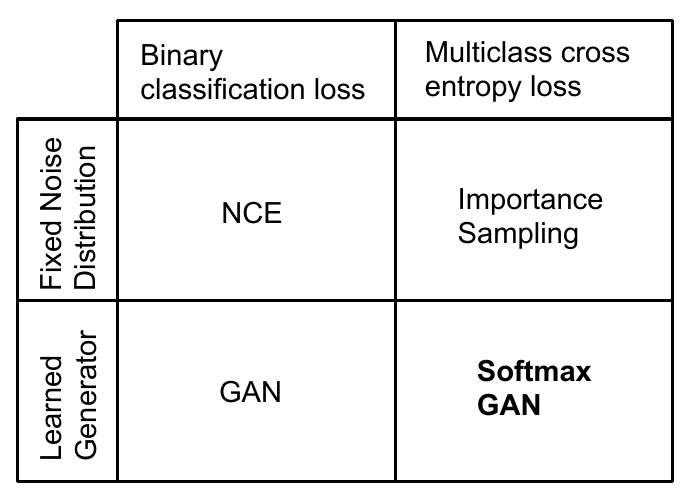}
\end{figure}

Note that Importance Sampling itself is strongly connected to NCE. As
pointed out by \cite{Link_NCE_IS} and \cite{Link_NCE_IS_web}, both
Importance Sampling and NCE are training the underlying generative
model with a classification surrogate. The difference is that in NCE,
a binary classification task is defined between true and noise samples
with a logistic loss, whereas Importance Sampling replaces the
logistic loss with a multiclass softmax and cross-entropy loss. We
show the relationship between NCE, Importance Sampling, GAN and this
work in Figure \ref{fig:comparechart}. Softmax GAN is filling the
table with the missing item.

\subsection{Infinite batch size}
As pointed out by \cite{bengio2008adaptive}, biased Importance
Sampling estimation converges to the real partition function when
the number of samples in one batch goes to infinity. In practice, we
found that setting $|B_+|=|B_-|=5$ is enough for generating images
that are visually realistic.

\section{Experiments}
We run experiments on image generation with the celebA database. We
show that although Softmax GAN is minimizing the \textit{Jensen
  Shannon divergence} between the generated data and the real data, it
is more stable than the original GAN, and is less prone to mode
collapsing.

We implement Softmax GAN by modifying the loss function of the DCGAN
code (https://github.com/carpedm20/DCGAN-tensorflow). As DCGAN is
quite stable, we remove the empirical techniques applied in DCGAN and
observe instability in the training. On the contrary, Softmax GAN is
stable to these changes.

\subsection{Stablized training}
\begin{figure}
  \centering
  \includegraphics[width=0.4\textwidth]{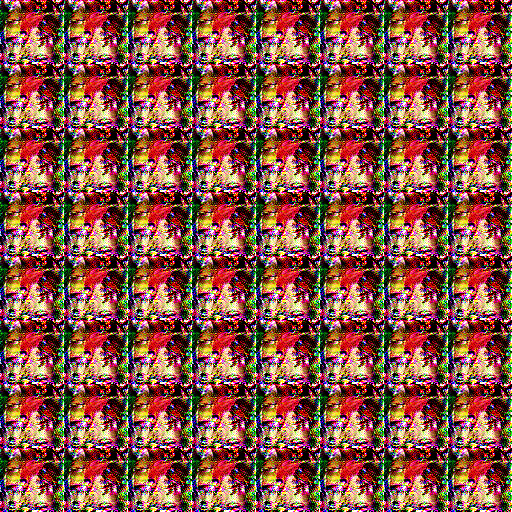}
  \includegraphics[width=0.4\textwidth]{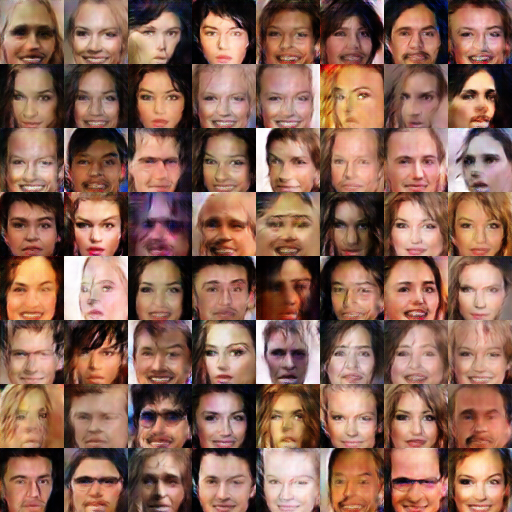}
  \caption{Generator without batch normalization and with a constant
    number of filters at each layer. Both GAN and Softmax GAN are able
    to train at the early stages. However, the discriminator loss of
    GAN suddenly drops to zero in the 7th epoch, and the generator
    generates random patterns (left). Softmax GAN (right) is not
    affected except that the generated images are of slightly lower
    qualities, which could be due to the reduced number of parameters
    in the generator.}
  \label{fig:unstable}
\end{figure}

We follow the WGAN paper, by removing the batch normalization layers
and using a constant number of filters in the generator network. We
compare the results from GAN and Softmax GAN. The results are shown in
Figure \ref{fig:unstable}.

\subsection{Mode collapse}
When GAN training is not stable, the generator may generate samples
similar to each other. This lack of diversity is called mode collapse
because the generator can be seen as sampling only from some of the
modes of the data distribution.

\begin{figure}
  \centering
  \includegraphics[width=0.4\textwidth]{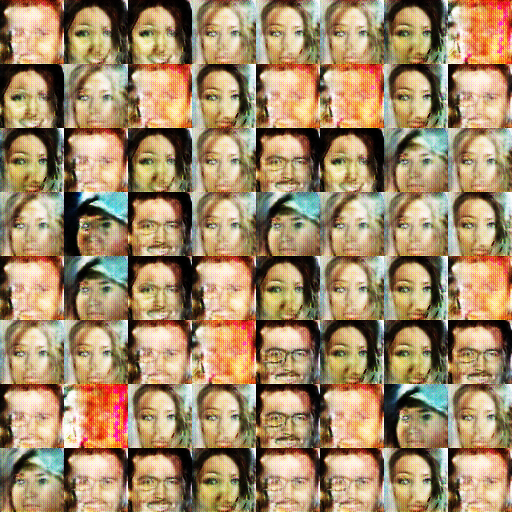}
  \includegraphics[width=0.4\textwidth]{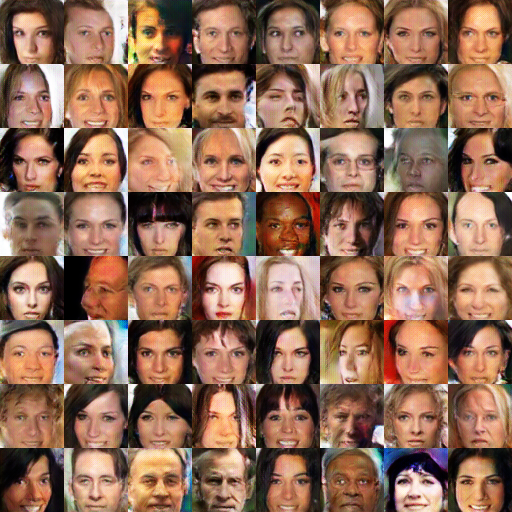}
  \caption{Using relu instead of leaky relu, and normalizing input to
    $[0, 1)$, GAN (left) suffers from mode collapse and low image quality,
      while Softmax GAN (right) is not affected.}
  \label{fig:modecollapse}
\end{figure}

In the DCGAN paper, the pixels of the input images are normalized to
$[-1, 1)$. We remove this constraint and instead normalize the pixels
  to $[0, 1)$. At the same time, we replace the leaky relu with relu,
    which makes the gradients more sparse (this is unfavorable for GAN
    training according to the DCGAN authors). Under this setting, the
    original GAN suffers from a significant degree of mode collape and
    low image qualities. In contrast, Softmax GAN is robust to this
    change. Examplars are show in Figure \ref{fig:modecollapse}.

\subsection{Balance of generator and discriminator}
\begin{figure}
  \centering
  \includegraphics[width=0.4\textwidth]{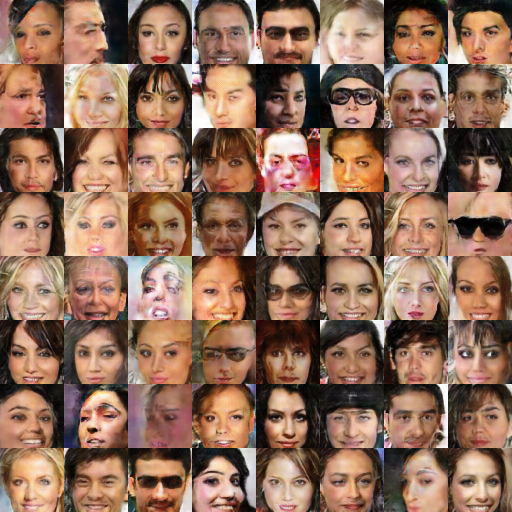}
  \includegraphics[width=0.4\textwidth]{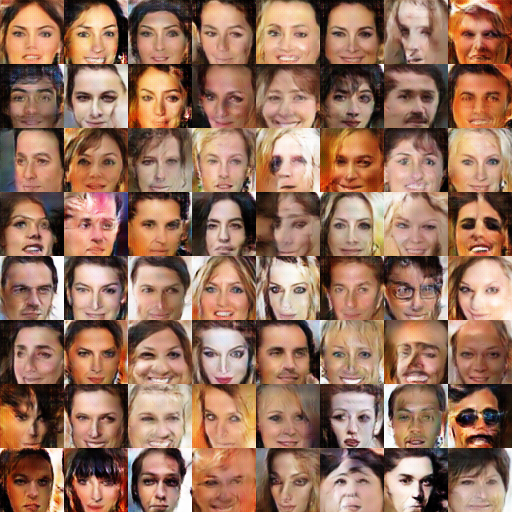}
  \caption{Training DCGAN (left) and Softmax GAN (right) with
    $\#discriminator\;iter / \#generator\;iter = 5 / 1$. The visual
    appearances of both GANs are similar.}
  \label{fig:51}
\end{figure}

\begin{figure}
  \centering
  \includegraphics[width=0.4\textwidth]{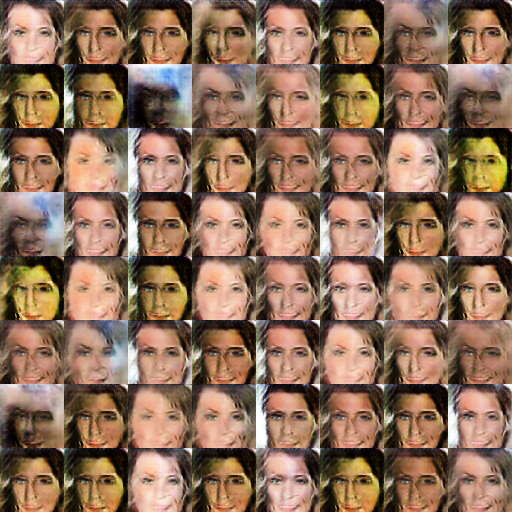}
  \includegraphics[width=0.4\textwidth]{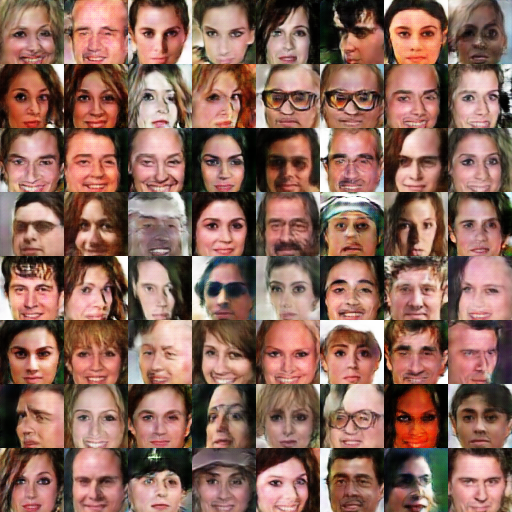}
  \caption{Training DCGAN and Softmax GAN with $\#discriminator\;iter
    / \#generator\;iter = 1 / 5$. The images from DCGAN generator
    become blurred and suffer from mode collapse (left). Softmax GAN
    is less affected (right).}
  \label{fig:15}
\end{figure}

It is claimed in \cite{DCGAN} that manually balancing the number of
iterations of the discriminator and generator is a bad idea. The WGAN
paper however gives the discriminator phase more iterations to get a
better discriminator for the training of the generator. We set the
discriminator vs generator ratio to $5:1$ and $1:5$ and explore the
effects of the this ratio on DCGAN and Softmax GAN. The results are in
Figure \ref{fig:51} and \ref{fig:15} respectively.

\section{Conclusions and future work}

We propose a variant of GAN which does softmax across samples in a
minibatch, and uses cross-entropy loss for both discriminator and
generator. The target is to assign all probability to real data for
discriminator and to assign probability equally to all samples for the
generator. We proved that this objective approximately optimizes the
JS-divergence using Importance Sampling. We futhur form the links
between GAN, NCE, Importance Sampling and Softmax GAN. We demonstrate
with experiments that Softmax GAN consistently gets good results when
GAN fails at the removal of empirical techniques.

In our future work, we'll perform a more systematic comparison between
Softmax GAN and other GAN variants and verify whether it works on
tasks other than image generation.

\bibliography{nips_2017}
\bibliographystyle{plain}

\end{document}